\pdfoutput=1

\documentclass[11pt]{article}

\usepackage[preprint]{acl}

\usepackage{times}
\usepackage{latexsym}

\usepackage{bm}
\usepackage{multirow}
\usepackage{booktabs}
\usepackage{graphicx}
\usepackage{subfig}
\usepackage{booktabs}
\usepackage{comment}
\usepackage{siunitx}
\usepackage{caption}
\usepackage[T1]{fontenc}

\usepackage[utf8]{inputenc}

\usepackage{microtype}

\usepackage{inconsolata}

\title{Can AI Extract Antecedent Factors of Human Trust in AI?}

\author{Melanie McGrath, Harrison Bailey, Necva Bölücü\\ 
\textbf{Xiang Dai, Sarvnaz Karimi, Cecile Paris}\\
  CSIRO Data61\\
  Australia\\
  \texttt{firstname.lastname@csiro.au} \\
}

\begin{document}
\maketitle
\begin{abstract}
Information extraction from the scientific literature is one of the main techniques to transform unstructured knowledge hidden in the text into structured data which can then be used for decision-making in down-stream tasks. One such area is {\em Trust in AI}, where factors contributing to human trust in artificial intelligence applications are studied. The relationships of these factors with human trust in such applications are complex. We hence explore this space from the lens of information extraction where, with the input of domain experts, we carefully design annotation guidelines, create the first annotated English dataset in this domain, investigate an LLM-guided annotation, and benchmark it with state-of-the-art methods using large language models in named entity and relation extraction. Our results indicate that this problem requires supervised learning which may not be currently feasible with prompt-based LLMs. 

\end{abstract}

\section{Introduction}\label{sec:intro}

The rapid rate at which \emph{Artificial Intelligence} (AI) is developing and the accelerating rate at which it is becoming integrated into human life necessitates a thorough understanding of the dynamics of human trust in AI~\citep{glikson2020human,teaming2022state}. Addressing questions about the factors, or \textit{antecedents}, influencing trust in specific AI systems and the thresholds for excessive or insufficient trust is crucial for using AI responsibly and preventing potential misuse~\citep{parasuraman_humans_1997,lockey_review_2021}.

Existing literature in the behavioural and computer sciences offers extensive insight into this domain~\cite[e.g.,][]{glikson2020human,kaplan_trust_2021, sassmannshausen2023human}. However, this literature remains largely unstructured and is constantly expanding, making it increasingly difficult for researchers to review and extract relevant knowledge. To address this challenge, we create the \texttt{Trust in AI} dataset where the factors influencing trust are automatically captured in a structured dataset, making it more accessible and easier for domain experts to navigate. The specific information targeted for extraction, as directed by the domain experts, includes the type of AI {\em application}, the {\em factor} (trust antecedent), the type of factor ({\em human}, {\em technological}, {\em contextual}), and the relationship of the factor with \textit{trust}. The resulting resource has practical applications in industry and commercial AI production. To build this resource, we utilise Information Extraction (IE) methods, including Named Entity Recognition (NER) and Relation Extraction (RE).

Our contributions are as follows: (1) We formulate the challenging problem of information extraction for trust in AI, which is previously unexplored in the NLP domain (\S\ref{sec:problem}); (2) Drawing inspiration from studies demonstrating the capabilities of large language models (LLMs) in simulating annotation~\citep{bansal2023large,goel2023llms, zhang2023llmaaa}, we provide LLM-guided annotation as a part of the annotation process; (3) We construct a dataset of factors empirically shown to influence trust development named \texttt{Trust in AI} (\S\ref{sec:dataset}); 
and, (4) We provide baseline results for the defined tasks (\S\ref{sec:results}).

\section{Related Work}\label{sec:related}
\paragraph{Trust in AI}
Trust is critical to human willingness to adopt AI technology in a safe and productive way~\citep{jacovi2021formalizing, schaefer_roadmap_2021}. Consequently, it is important to know what factors contribute to the development of an appropriate level of trust in an AI application. Over $450$ distinct factors influencing trust development have been identified in the scientific literature~\citep{sassmannshausen2023human}. These antecedent factors can be classified as (1) properties of the trustor or {\em human} factors (e.g., experience); (2) properties of the trustee or {\em technological} factors (e.g., performance); or (3) properties of the task or interaction {\em context} (e.g., time pressure)~\citep{hancock_meta-analysis_2011, kaplan_trust_2021, schaefer_meta-analysis_2016}. 

Which of these hundreds of antecedents influence trust in a particular AI application is highly variable. As a result, researchers interested in trust development are increasingly seeking approaches to specifying idiosyncratic models of trust in individual applications. The \texttt{Trust in AI} dataset will provide these domain experts with a resource to identify the most relevant factors for their application based on the existing literature. To our knowledge, this is the first such resource created for use by both NLP and Trust in AI researchers.

\paragraph{Scientific IE Dataset Annotation}
Annotating scientific IE datasets can be approached in two key ways: (a) annotating a small amount of data with the help of domain experts and carefully designed annotation guidelines~\citep{friedrich-bosch-2020-acl-sofc-exp,karimi-csiro-2015-jbi-cadec,kim-jst-2003-bioinformatics-genia}; and, (b) leveraging existing resources/tools to automatically annotate a large amount of data with no or little human intervention~\citep{agrawal-etal-2019-scalable,jain-etal-2020-scirex}.
 
Each group of studies have their pros and cons with the trade-off of cost, scale, and precision in annotations. Our study fits in the first category as the concepts of interest and their relationships are complex which entails an expert annotation for this first attempt to create such a resource in trust in AI domain. 

\paragraph{IE using LLMs}
IE using LLMs has gained prominence in the literature due to its potential advantages, particularly in scenarios with limited annotated data or in domains where traditional supervised approaches face challenges~\citep{brown2020language, bubeck2023sparks}. LLMs show the capability of few-shot learning, useful when annotated data is scarce or expensive to obtain~\citep{agrawal2022large, li2023codeie, zhang2023aligning}. By employing prompt techniques, LLMs provide a consistent approach to various IE tasks through a single model~\citep{wang2022deepstruct}. However, utilising LLMs with instructions for IE has been less successful~\citep{bolucu2023impact, gutierrez2022thinking, zhou2023universalner, zhang2023promptner}. There is also a surge of generative IE methods that leverage LLMs to generate structural information rather than extracting structural information from text~\citep{guo2023retrieval, sainz2023gollie,qi2023preserving}.

\section{Problem Formulation}\label{sec:problem}
The \texttt{Trust in AI} dataset, can be conceptualised as $D = \{S_i, P_i, L_i, R_i\}_{i=1}^N$, where $N$ is the total number of sentences in the dataset. For each sentence $S_i$, $P_i$ represents its context, which is the paragraph where the sentence $S_i$ is located, $L_i$ is the set of entity mentions, and $R_i$ is the set of identified relations within the sentence.
Each element (entity) in $L_i$ is represented as a triplet, consisting of the start index of a span, the end index, and the entity category (i.e., human factor, technology factor, context factor, and application name).
Each element in $R_i$ captures the relationship between one or a combination of two factors and the hidden concept \textit{trust}.
Therefore, depending on the number of factors involved, elements in $r_i$ are represented by either a tuple $(L_i^m, r)$ or a triplet $(L_i^m, L_i^n, r)$, where $r$ is one of the pre-defined relation type: {\em unspecified}, {\em null}, {\em positive}, {\em negative}, {\em interaction}, or {\em no relation}.

The dataset can be used for benchmarking two tasks: (1) {\em Named Entity Recognition}; for a given sentence $S_i$ and the related paragraph $P_i$, the objective is to recognise all elements in $L_i$; and, (2) {\em Relation Extraction}; for a given sentence $S_i$, the related paragraph $P_i$ and the recognised entities $L_i$, the objective is to extract all elements in $R_i$.\footnote{It is worth noting that the relation we investigate is between entities mentioned in the text and the hidden concept \textit{trust}, in contrast to the common relation classification task, where relation exists between two entity mentions.}

\section{Trust in AI Dataset}\label{sec:dataset}

Our dataset comprises a diverse collection of English scientific publications focused on \textit{trust in automation and AI} and \textit{trust in collaboration with AI}. Appendix~\ref{ssec:search} provides details on the dataset curation.

Two annotators (one researcher holding a PhD in social psychology and one student majoring in computer science and politics) take part in the annotation task. The annotation unfolded in two stages. In {\em Stage I}, the focus is on annotating factors and applications as spans, while {\em Stage II} focuses on annotating relationships between factors and the concept of \textit{trust}. The Prodigy annotation tool~\citep{montani2018prodigy} is employed for both stages, with details about the annotation interface available in Appendix~\ref{ssec:interface}. The annotation of each stage is done in $5$ phases: 
\paragraph{\romannumeral 1. Preparation of the guideline:} The annotation guideline is developed through a small pilot annotation by one annotator using $5$ publications.
\paragraph{\romannumeral 2. Annotation:} All annotators annotate the same $5$ publications used in the first phase.
\paragraph{\romannumeral 3. Resolution:} Annotators discuss their annotations, leading to updates in the annotation guideline to address any errors or discrepancies.
\paragraph{\romannumeral 4. LLM-guided annotation:} Inspired by studies in dataset annotation~\citep{bansal2023large,goel2023llms, zhang2023llmaaa}, we utilise LLMs as guidance for annotators during the annotation of $6$ randomly allocated publications per annotator, resulting in a total of $12$ publications. We display the predictions of LLMs (details in Appendix~\ref{ssec:llm_guidence}) to assist annotators in the process and reduce annotation time. Annotators rectify any errors made by LLM. This allows us to compare the effectiveness of LLM with manual annotation. 
\paragraph{\romannumeral 5. Manual annotation:} The remaining articles are annotated using the updated guidelines.

\paragraph*{Inter-Annotator Agreement}
After phase $ii$, inter-annotator Cohen's kappa ($\kappa$) scores~\citep{cohen1960coefficient} are $39.5$ and $32.3$ for Stage I (NER) and Stage II (RE), respectively. Upon resolution in phase $iii$, it is observed that, for NER, the main disagreement is the annotation of {\em application} and {\em technology}. In contrast, for RE, it is mainly related to the annotation of {\em interaction}. Following resolution, substantial overall agreements of $93.3$ (NER) and $93.5$ (RE) are achieved. Kappa values over $90$ are considered near perfect agreement~\citep{cohen1960coefficient,mchugh2012interrater}, possibly reflecting the high coverage of annotation guidelines and training of the annotators. Additionally, as annotators use the same dataset at Stage I (NER) and Stage II (RE), familiarity with the articles may have assisted them in identifying relationships in Stage II.

The Cohen's Kappa scores between LLM-agent and human annotators are low ($12.85$ for NER and $5.59$ for RE), highlighting the complexity of these tasks for LLMs and the necessity of a human annotated dataset.

\begin{table}[tb]
\resizebox{\columnwidth}{!}{
    \begin{tabular}{r rrr r}
    \toprule
    \textbf{Statistic} &  \textbf{Train} &  \textbf{Dev} &  \textbf{Test} &  \textbf{Overall} \\
    \midrule
    \# documents & \multicolumn{2}{c}{--22--} & 5 & 27 \\
    \# paragraphs & 213 & 24 & 38 & 275 \\
    \# sentences & 1,249 & 126 & 233 & 1,612 \\
    \# tokens & 32,201 & 3,128 & 6,836 & 42,165 \\
    \midrule
    \# entities & 2,454 & 268 & 535 & 3,260 \\
    \# relations & 3,920 & 344 & 874 & 5,138 \\ 
    \bottomrule
    \end{tabular}
}
\caption{Descriptive statistics of \texttt{Trust in AI}.}\label{tab:dataset-statistic}
\end{table}

\begin{table*}[th]
\centering
\small
    \begin{tabular}{l cccc c cc}\toprule
     & \multicolumn{4}{c}{\textbf{Category}} && \multicolumn{2}{c}{\textbf{Overall}} \\\cline{2-5} \cline{7-8}
   \textbf{Method} & \textbf{Application} & \textbf{Human fac} & \textbf{Technology fac} & \textbf{Contextual fac} && \textbf{Micro F\textsubscript{1}} & \textbf{Macro F\textsubscript{1}} \\
    \midrule
    RoBERTa-Base & 84.5 \tiny $\pm$ 0.8 & 48.6 \tiny $\pm$ 7.8 & 54.6 \tiny $\pm$ 8.7 & 51.1 \tiny $\pm$ 4.2 && 60.9\tiny $\pm$ 3.5 & 55.1 \tiny $\pm$ 4.3\\
    RoBERTa-Large & \textbf{85.4} \tiny $\pm$ 1.5 & \textbf{50.1} \tiny $\pm$ 6.5 & \textbf{55.2} \tiny $\pm$ 7.9 & \textbf{53.1} \tiny $\pm$ 3.8 && \textbf{61.2} \tiny $\pm$ 3.1 & \textbf{57.2} \tiny $\pm$ 3.8 \\
    Seq2seq-BERT & 72.5\tiny $\pm$ 1.1 & 34.8\tiny $\pm$ 3.5 & 44.2 \tiny $\pm$ 4.5 & 45.7 \tiny $\pm$ 2.2 && 50.7 \tiny $\pm$ 2.1 & 46.2 \tiny $\pm$ 2.6\\
    BiaffineNER & 75.5 \tiny $\pm$ 0.9 & 39.2 \tiny $\pm$ 2.5 & 48.5 \tiny $\pm$ 3.2 & 49.2 \tiny $\pm$ 1.2 && 55.2 \tiny $\pm$ 1.5 & 50.5 \tiny $\pm$ 3.0\\
    Few-shot Learning & 26.1 \tiny $\pm$ 0.3 & 12.5 \tiny $\pm$ 0.6 & 18.1 \tiny $\pm$ 0.4 & \phantom{1}9.3 \tiny $\pm$ 0.2 && 18.3 \tiny $\pm$ 0.4 & 18.3 \tiny $\pm$ 0.3 \\\bottomrule
    \end{tabular}
    
\caption{Comparison of models in terms of entity-level F\textsubscript{1} for NER task. The best results are \textbf{boldfaced}. `fac' stands for factor. We report results of the best-performing selection method for Few-shot Learning, and the full results can be found in Appendix~\ref{sec:all}.}\label{tab:ner_results}
\end{table*}

\begin{table}[th]
\small
\setlength{\tabcolsep}{4pt}
    \centering
    \begin{tabular}{l c rr}
    \toprule
      & \textbf{Identification} & \multicolumn{2}{c}{\textbf{Classification}} \\
      \textbf{Method} & \textbf{ROC-AUC} & \textbf{Micro F\textsubscript{1}} & \textbf{Macro F\textsubscript{1}} \\
    \midrule
Random & 50.5 \tiny $\pm$ 1.7  & 34.1 \tiny $\pm$ 3.3 & 16.9 \tiny $\pm$ 3.7  \\ 
GPT-4 ($0$-shot) & 52.2 \tiny $\pm$ 0.2 & 52.3 \tiny $\pm$ 0.5 & 33.5 \tiny $\pm$ 0.4 \\ 
RoBERTa-Base & 79.8 \tiny $\pm$ 3.6 & 53.3 \tiny $\pm$ 1.2 & 45.6 \tiny $\pm$ 1.4  \\ 
RoBERTa-Large & \textbf{81.7} \tiny $\pm$ 0.2 & \textbf{57.7} \tiny $\pm$ 6.6 & \textbf{53.7} \tiny $\pm$ 11.5\\ 
    \bottomrule
    \end{tabular}
    \caption{Evaluation results of RE task. The best results are \textbf{boldfaced}.}
    \label{tab:rel_main_results}
\end{table}

\section{Experimental Setup
}\label{sec:ex_setup}
\paragraph*{Dataset}
We split our dataset into training, development, and test sets. The test set contains samples annotated by two annotators (phase $ii$ and $iii$), while the remaining sets are annotated by one annotator, both with (phase $iv$) and without LLM guidance (phase $v$). Descriptive statistics of the dataset are given in Table~\ref{tab:dataset-statistic}.

\paragraph{NER models} We benchmark the effectiveness of several models: (1) {\em RoBERTa}, which is composed of a RoBERTa encoder~\citep{liu2019roberta} and span-based classifier on top of the encoder~\citep{zhong2021frustratingly}; (2) {\em Seq2seq-BERT}\footnote{\url{https://github.com/ufal/acl2019_nested_ner}}~\citep{strakova2019neural}, a sequence-to-sequence model consisting of an encoder-decoder with LSTM; (3) {\em BiaffineNER}\footnote{\url{https://github.com/LindgeW/BiaffineNER}}~\citep{yu2020named}, formulated as a graph-based parsing task composed of a BiLSTM encoder with a biaffine classifier; and (4) {\em Few-shot learning} by leveraging in-context learning (ICL) methods as formulated by~\citet{bolucu2023impact}.

\paragraph{RE models} Since the RE annotations are highly imbalanced, we address the task in two steps: (1) relation identification, determining whether there is a relationship between a given factor and trust; and (2) relation type classification, determining the fine-grained relation type. For both subtasks, we compare the following methods: (1) {\em Random}, assigning labels to test examples according to the label distribution from the training set; (2) {\em RoBERTa}~\citep{liu2019roberta}, employing fine-tuned PLMs on the training data; and (3) {\em Zero-shot learning using GTP-4}, using prompts designed for the two subtasks (see Appendix~\ref{sec:ex_setup} for details).

\paragraph*{Evaluation Metrics}
We use entity-level F\textsubscript{1} score~\citep{seqeval} for NER, Area Under the Receiver Operating Characteristic Curve (ROC-AUC) for relation identification, and F\textsubscript{1} scores for relation classification. 
All experiments are repeated three times and mean values and standard deviations are reported. 

\section{Results and Analysis}\label{sec:results}
The results are shown in Table~\ref{tab:ner_results} for NER and~\ref{tab:rel_main_results} for RE. We observe that the supervised models outperform those using LLM, consistent with the study of~\citep{gutierrez2022thinking,bolucu2023impact}. This finding highlights the necessity of a human annotated dataset for this complex domain. 

At the annotation of factor types and application at Stage I, one word can refer to one or more factor types. For instance, the word \textit{adaptability} in \textit{user and robot adaptability} refers to both {\em human} and {\em technology} factors. Moreover, a mentioned factor may span several words, not all of which are included in the same factor. For instance, a publication might mention \textit{training of communication and trust calibration}, where \textit{training of communication} is a {\em technology} factor while \textit{training $\cdots$ of trust calibration} is {\em human} factor. This complexity makes the NER task challenging. Even though span-based models are applied to extract factors and applications, the results of SOTA models on the annotated dataset remain relatively low except for {\em application}. We note that {\em human} and {\em contextual} factors are the most confused factors. As a result, the annotation guideline is updated to provide clarity in distinguishing between the annotation of these factors. This shows that extracting these factors is even challenging for human annotators who have expertise in the domain. Finally, {\em application} is expected to be used to label entities that may contain the AI technology or the studied use case, potentially contributing to lower results for the {\em technology} factor.

The article selection, carried out by a domain expert, is crucial for the annotation of the relation between \textit{trust} and factors at the RE task. Even in articles analysing \textit{trust in AI}, factors may be discussed without any explicit relationship to trust being stated. This would be labelled as {\em no relation} in the dataset. The majority of factors in any given paper attract the {\em no relation} label, leading to an imbalanced distribution of relations in the dataset. The distribution of remaining relations is also imbalanced (see Table~\ref{tab:table_data_statistics} in Appendix~\ref{ssec:f_analyses} for details), with many labelled as {\em unspecified}, indicating that a relation is reported but its nature is not described. 
Hence the RE task is designed in two steps: first identifying the relation and then detecting the relation type. 
The supervised model outperforms LLM at both relation identification and relation classification.
We also find LLM performs much better at relation classification than identification, as it predicts a relationship exists between most of the factors and trust, resulting in low precision in the lower.

\section{Conclusion and Future Work}\label{sec:conclusion}
We explore information extraction for \textit{trust in AI} and construct a dataset, comprising scientific articles. The dataset is constructed by annotating {\em application} and factors influencing \textit{trust}, and the relation between \textit{trust} and factors, formulated as named entity recognition and relation extraction tasks. We benchmark these tasks showing that LLM models using zero- and few-shot learning underperform supervised models, highlighting the need for the annotated dataset for this domain.

In the future, we plan to extend the dataset for entity resolution to identify and link entities that refer to the same entity, providing a more cohesive and accurate representation.

\section*{Ethics Statement}
As we create a dataset, there are ethical considerations about using the data. The dataset used in our work is collected from scientific publications which are publicly available. However, some may require subscriptions to the journals for their users. We make links to the articles available so as not to redistribute those without their publishers' permission.

\section*{Limitations}
{\em Language.} This dataset only uses English scientific literature, which may limit its usage for other languages. 

{\em Subjectivity and Background Knowledge.} The dataset annotation is done by two human annotators with different background knowledge, with one expert in the \textit{Trust in AI} domain with a psychology background and another in computer science and politics. 

{\em Access.} Due to IP considerations, the annotated dataset will be provided to the first author on requests.

\section*{Acknowledgements}
Harrison Bailey was an intern at CSIRO's Data61 from November 2023 to February 2024, where he contributed to this research.

\bibliography{custom}

\appendix

\section{Details of Dataset}\label{sec:details}

\subsection{Dataset Curation}\label{ssec:search}
The publications are provided by domain experts, following a systematic review. To search the literature and gather scientific articles, $2$ queries with terms related to \textit{artificial intelligence}, \textit{robotics}, \textit{automation}, and \textit{trust-related} concepts are used: 

\paragraph{Trust in automation and AI query: }
\hfill \break
({\em artificial intelligence} \textbf{OR} {\em robot*} \textbf{OR} {\em automation} \textbf{OR} {\em machine intelligence} \textbf{OR} {\em autonomy}) \\
    \textbf{AND} \\
    ({\em trust*} \textbf{OR} {\em trust models} \textbf{OR} {\em trustworthiness} \textbf{OR} {\em trust calibration} \textbf{OR} {\em trust repair} \textbf{OR} {\em trust propensity} \textbf{OR} {\em trust development}) 
   
\paragraph{Trust in collaboration with AI query: }
\hfill \break
({\em human-robot collaboration} \textbf{OR} {\em hybrid intelligence} \textbf{OR} {\em collaborative intelligence} \textbf{OR} {\em robot*} \textbf{OR} ({\em collaboration} \textbf{AND} {\em artificial intelligence}) \textbf{OR} {\em human-AI collaboration} \textbf{OR} {\em human-robot team*} \textbf{OR} {\em human-autonomy team*} \textbf{OR} {\em augmented intelligence} \textbf{OR} {\em human-machine team*})\\
    \textbf{AND}\\
    ({\em trust*} \textbf{OR} {\em trust models} \textbf{OR} trustworthiness\textbf{OR} {\em trust calibration} \textbf{OR} {\em trust repair} \textbf{OR} {\em trust propensity} \textbf{OR} {\em trust development})

After conducting searches, a domain expert screens publications for relevance to the literature. The criteria for inclusion are addressing the topics; \textit{definitions of trust}, \textit{experimental manipulations of trust}, \textit{measurement of trust}, \textit{antecedents of trust}, or \textit{outcomes of trust}.  From the full set of papers retained for the review, a subset of publications coded as referencing antecedents of trust is used to construct the initial \texttt{Trust in AI} dataset. 

As the purpose of the \texttt{Trust in AI} dataset is to capture trust antecedents that have been empirically tested, annotations are confined to those sections of a paper most likely to report its own findings, namely the abstract, results, and conclusion. Given that papers are collected from several disciplines, including robotics, psychology, human factors, ergonomics, and system sciences, the titles used to denote these sections of the paper vary. To address this, one annotator with knowledge of the domain reviews each paper in the dataset to identify the headings corresponding to the abstract, results, or conclusions sections.  

\subsection{Annotation Interface}
\label{ssec:interface}
\begin{figure*}[tb]
   \centering
    \includegraphics[width=1.0\textwidth]{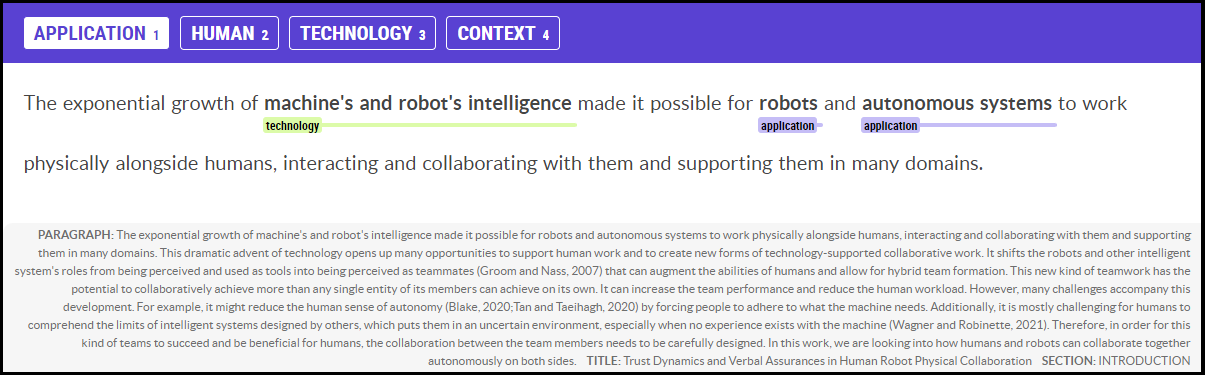}
    \caption{Interface for manual annotation of the NER task.}
    \label{fig:interface}
\end{figure*}

The Prodigy annotation tool is utilised for annotation. We design a web page that integrates the Prodigy annotation tool, allowing annotators to input their names and select the publication and specific sections before initiating the annotation process. The interface for NER is illustrated in Figure~\ref{fig:interface}. As depicted in the figure, context information, including the paragraph containing the sentence, as well as the title and section names of the publication, is presented for each sentence.

\begin{figure}[tb]   \centering
    \includegraphics[width=0.47\textwidth]{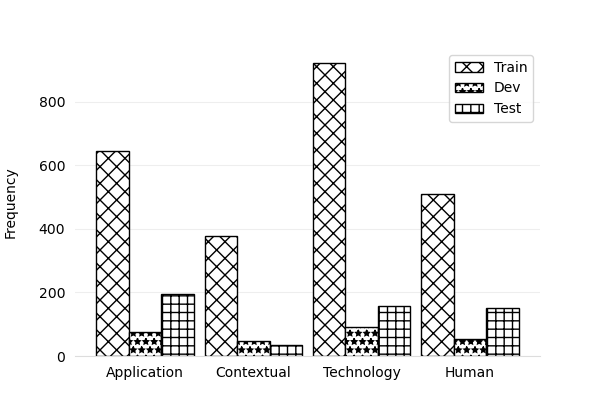}
    \caption{Distribution of factors and application of our annotated dataset \texttt{Trust in AI}.}
    \label{fig:distribution}
\end{figure}

\subsection{LLM Guidance Annotation}\label{ssec:llm_guidence}
For LLM guidance, we called the \texttt{gpt-3.5-turbo} version of ChatGPT integrated into the spaCy library\footnote{\url{https://spacy.io/}}~\citep{honnibal2020spacy}. We adopted a temperature of $0.3$ and used ICL (random sampling- $3$-shot) with \texttt{spacy.SpanCat.v2} and \texttt{spacy.TextCat.v2} components of spaCy and prompt defined by the library to pre-annotate the dataset for NER and RE tasks, respectively.

\subsection{Further Analyses}\label{ssec:f_analyses}
The distribution of application and factor types (contextual, technology, human) annotated in NER (Stage I) is presented in Figure~\ref{fig:distribution}.

As stated, we examine at most two factors influencing the relation with \textit{trust} (see Section~\ref{sec:problem} for details). The statistics of the factors are presented in Table~\ref{tab:table_data_statistics} to provide insight into the task.

\begin{table*}[tb]
    \centering
    \begin{tabular}{r rrr }
    \toprule
         &  Train & Dev & Test \\
    \midrule
\# Examples & 3920 & 344 & 874 \\ 
\# Examples w. relation & 713 & 92 & 100 \\ 
Average sentence length & 38.6 & 33.3 & 48.2 \\ 
    \hline
\# Single-factor Examples & 1769 & 204 & 377 \\ 
\# Single-factor Examples w. relation & 622 (465/43/93/17/4) & 74 (67/-/6/1/-) & 91 (42/20/29/-/-)\\ 
    \hline
\# Double-factor Examples & 2151 & 140 & 497 \\ 
\# Double-factor Examples w. relation & 91 (6/3/-/1/81) & 18 (-/-/-/-/18) & 9 (-/-/-/-/9) \\ 
    \bottomrule
    \end{tabular}
    \caption{Relation extraction annotation statistics giving single- and double-factor examples separately. The distribution of relation types in brackets in order of {\em Unspecified}, {\em Null}, {\em Positive}, {\em Negative}, and {\em Interaction} for single- and double-factors for each set.}
    \label{tab:table_data_statistics}
\end{table*}

\section{Experimental Setups}\label{sec:ex_setup}
All hyper-parameters used in supervised baseline models for NER and RE tasks are tuned on the development set. The details of experiments of SOTA models are given below for NER and RE tasks.

\paragraph*{NER}
For RoBERTa, the hyperparameters are the learning rate of 5e-4, max length of $128$, context window of $200$ tokens and batch size of $16$, and models are trained for max epochs of $30$. For Seq2seq-BERT and BiaffineNER, we use the default hyperparameters suggested by the authors except for the learning rate which is 5e-3.

For zero- and few-shot learning, we adopt the prompt template provided by EasyInstruct library\footnote{\url{https://github.com/zjunlp/EasyInstruct}}~\citep{ou2024easyinstruct} for ICL and use GPT3.5 and set the temperature to $0.1$ to get results with less variability. We follow the study of~\citep{bolucu2023impact} to select ICL samples. The prompt used for NER is:

\begin{table*}[!ht]
\small
\centering
   \begin{tabular}{l cccc c cc}\toprule
     & \multicolumn{4}{c}{\textbf{Category}} && \multicolumn{2}{c}{\textbf{Overall}} \\\cline{2-5} \cline{7-8}
   \textbf{Method} & \textbf{Application} & \textbf{Human fac} & \textbf{Technology fac} & \textbf{Contextual fac} && \textbf{Micro F\textsubscript{1}} & \textbf{Macro F\textsubscript{1}} \\
    \midrule
    $0$-shot  & 11.4 \tiny $\pm$ 0.1 & 10.5 \tiny $\pm$ 0.2 & 6.8 \tiny $\pm$ 0.0 & 7.0 \tiny $\pm$ 0.2 && 8.0 \tiny $\pm$ 0.5 & 9.0 \tiny $\pm$ 0.3\\\hline
    \multicolumn{7}{l}{\textit{Random Sampling}}\\
    $1$-shot & 14.5 \tiny $\pm$ 1.2 & 11.2 \tiny $\pm$ 0.9 & 11.6 \tiny $\pm$ 1.1 & 9.5 \tiny $\pm$ 0.9 && 13.3 \tiny $\pm$ 1.4 & 13.5 \tiny $\pm$  1.0\\
    $3$-shot & 18.4 \tiny $\pm$ 1.5 & 9.2 \tiny $\pm$ 1.3 & 9.2 \tiny $\pm$ 0.9 & 9.1 \tiny $\pm$ 0.7 && 12.4 \tiny $\pm$ 1.7 & 12.4 \tiny $\pm$ 1.5 \\
    $5$-shot & 20.1 \tiny $\pm$ 2.3 & 10.2 \tiny $\pm$ 1.6 & 13.3 \tiny $\pm$ 1.9 & \textbf{11.5} \tiny $\pm$ 2.1 && 14.2 \tiny $\pm$ 2.8 & 14.5 \tiny $\pm$ 2.2 \\\hline
    \multicolumn{7}{l}{\textit{BM25}}\\
    $1$-shot & 16.2 \tiny $\pm$ 0.1 & 7.1 \tiny $\pm$ 0.2 & 12.4 \tiny $\pm$ 0.3 & 9.3 \tiny $\pm$ 0.0 && 11.0 \tiny $\pm$ 0.1 & 11.0 \tiny $\pm$ 0.1 \\
    $3$-shot & 24.8 \tiny $\pm$ 0.4 & 8.6 \tiny $\pm$ 0.2 & 19.2 \tiny $\pm$ 0.4 & 9.1 \tiny $\pm$ 0.2 && 15.2 \tiny $\pm$ 0.2 & 14.8 \tiny $\pm$ 0.3\\
    $5$-shot & \textbf{26.1} \tiny $\pm$ 0.3 & 12.5 \tiny $\pm$ 0.6 & 18.1 \tiny $\pm$ 0.4 & 9.3 \tiny $\pm$ 0.2 && \textbf{18.3} \tiny $\pm$ 0.4 & \textbf{18.3} \tiny $\pm$ 0.3\\\hline
    \multicolumn{7}{l}{\textit{KATE - (cosine w/ max)}}\\
    $1$-shot  & 19.3 \tiny $\pm$ 0.7 & 8.3 \tiny $\pm$ 0.4 & 13.5 \tiny $\pm$ 0.4 & 9.4 \tiny $\pm$ 0.5 && 14.2 \tiny $\pm$ 0.6 & 13.6 \tiny $\pm$ 0.5 \\
    $3$-shot & 20.3 \tiny $\pm$ 0.4 & 9.4 \tiny $\pm$ 0.4 & 18.6 \tiny $\pm$ 0.4 & 10.6 \tiny $\pm$ 0.1 && 15.8 \tiny $\pm$ 0.4 & 15.2 \tiny $\pm$ 0.3\\
    $5$-shot & 24.2 \tiny $\pm$ 0.8 & \textbf{17.5} \tiny $\pm$ 0.3 & \textbf{19.8} \tiny $\pm$ 0.3 & 10.0 \tiny $\pm$ 0.3 && 18.2 \tiny $\pm$ 0.6 & 17.5 \tiny $\pm$ 0.5\\\hline
    \multicolumn{7}{l}{\textit{Perplexity}}\\
    $1$-shot  & 9.3 \tiny $\pm$ 0.6 & 7.7 \tiny $\pm$ 0.3 & 8.7 \tiny $\pm$ 0.2 & 7.8 \tiny $\pm$ 0.2 && 8.0 \tiny $\pm$ 0.2 & 8.3 \tiny $\pm$ 0.2\\
    $3$-shot & 11.0 \tiny $\pm$ 0.4 & 10.3 \tiny $\pm$ 0.2 & 14.1 \tiny $\pm$ 0.2 & 5.9 \tiny $\pm$ 0.0 && 10.2 \tiny $\pm$ 0.1 & 10.3 \tiny $\pm$ 0.2\\
    $5$-shot  & 11.2 \tiny $\pm$ 0.4 & 8.9 \tiny $\pm$ 0.1 & 10.3 \tiny $\pm$ 0.2 & 7.6 \tiny $\pm$ 0.0 && 9.3 \tiny $\pm$ 0.2 & 9.2 \tiny $\pm$ 0.3\\
    \bottomrule
    \end{tabular}
\caption{$0$, $1$, $3$, $5$ shot results of each ICL method for NER task. The best results are \textbf{boldfaced}. `fac' for factor.}\label{tab:ICL-full-results-NER}
\end{table*}

\begin{table*}[tb]
    \centering
    \begin{tabular}{r rrr}
    \toprule
    & \multicolumn{3}{c}{\textbf{Factor Category}}\\\cline{2-4}
      \textbf{Method}   & \textbf{Human} & \textbf{Technology} & \textbf{Contextual} \\
    \midrule
Random & 30.0 \tiny $\pm$ 3.8 & 38.8 \tiny $\pm$ 3.2 & 23.8 \tiny $\pm$ 7.3 \\ 
GPT-4 & 67.9 \tiny $\pm$ 1.8 & 39.1 \tiny $\pm$ 0.8 & 75.0 \tiny $\pm$ 0.0 \\ 
RoBERTa-Base & 39.7 \tiny $\pm$ 3.6 & 66.1 \tiny $\pm$ 0.8 & 29.2 \tiny $\pm$ 2.9 \\ 
RoBERTa-Large & 53.8 \tiny $\pm$ 19.1 & 64.4 \tiny $\pm$ 2.9 & 39.6 \tiny $\pm$ 20.6 \\
    \bottomrule
    \end{tabular}
    \caption{Factor-wise $F_1$ scores of relation classification.}
    \label{tab:table_factor_wise_relation_resuls}
\end{table*}

{\em \#\#\# Instruction: You are a highly intelligent and accurate span-based Named Entity Recognition (NER) system. The domain in which you complete this task is [the scientific literature concerning trust in AI]. You take Text as input and your task is to recognize and extract specific types of named entities in that given text and classify them into a set of predefined entity types: application, contextual, technology, and human.}

{\em application: This entity refers to parts of the text that specify the use case of the AI/collaborative task being studied.}

{\em human: This entity refers to parts of the text that identify, describe or refer to a trust antecedent (or factor) studied in the article which is a property of the human/trustor using the AI.}

{\em technology: This entity refers to parts of the text that identify, describe or refer to a trust antecedent (or factor) studied in the article which is a property of the AI/trustee being used.}

{\em context: This entity refers to parts of the text that identify, describe or refer to a trust antecedent (or factor) studied in the article which is a property of the task/interaction between the human and the AI, or a property of the environment in which the task/interaction takes place.}

{\em \#\#\# Context: Here is the sentence I need to label: [sentence]}

\paragraph*{RE}
As the input of RoBERTa, we insert special markers, i.e., $\&$ and $\#$, before and after the first and second target factors, respectively. Then, the concatenation of contextual token representations corresponding to these special markers is taken as input to the classification layer. For relation identification, the hyperparameters are the max length of $512$, batch size of $32$ and max epochs of $30$. For fine-tuning RoBERTa-Base, we use a learning rate of 5e-5, and for RoBERTa-Large, we use 2e-5. We use the same batch size and max length for relation type detection. The learning rate is set to 2e-5 for RoBERTa-Base with $10$ epochs, while for RoBERTa-Large, we use a learning rate of 5e-5 and $30$ epochs.

For zero- and few-shot learning experiments, we use GPT-4 and set the temperature to $0.0$. We use two prompts designed for two steps: 

\begin{enumerate}
    \item \textbf{Identification of relation:} The context is provided with and without a sentence for the task.

    {\em\#\#\# Instruction: You are an expert psychologist studying the factors that influence trust in AI. Your task is to determine whether there is a relationship between a given factor and trust, based on the provided sentence. Your response must be yes or no.}

    {\em\#\#\# Context: [sentence] Is there a relationship between the factor(s) [factor] and trust?}

    \item \textbf{Relation type detection:} The context is provided with and without a sentence for the task.

    {\em\#\#\# Instruction: You are an expert psychologist studying the factors that influence trust in AI. Your task is to determine the relationship between a given factor and trust, based on the provided sentence. Your response must fall into one of the following categories: positive effect, negative effect, no statistically significant empirical effect, exists but unspecified.}

    {\em\#\#\# Context: [sentence] What is the relationship between [factor(s)] and trust?}
\end{enumerate}

\section{All results}\label{sec:all}
\paragraph*{NER}
We applied the ICL methods explained in the study of~\citet{bolucu2023impact}. The results of each ICL sample selection method are given in Table~\ref{tab:ICL-full-results-NER}.

\paragraph*{RE}\label{pr:rel_identification}

For RE, the results of the RE task based on factor type are presented in Table~\ref{tab:table_factor_wise_relation_resuls}. 
\end{document}